\documentclass[review]{elsarticle}

\usepackage{lineno,hyperref}

\journal{arXiv}
\usepackage{amsmath,amsfonts}
\usepackage{amssymb}
\usepackage{latexsym}
\usepackage{algorithmic}
\usepackage{graphicx}
\usepackage{textcomp}
\usepackage{xcolor}
\usepackage{caption}
\usepackage{subcaption}
\usepackage{booktabs}
\usepackage[T1]{fontenc}
\usepackage{amsmath}
\usepackage{balance}
\usepackage{xurl,multirow}
\usepackage{soul}
\usepackage{microtype}
\usepackage[linesnumbered,ruled,vlined,noline,noend]{algorithm2e}

\usepackage{url}
\definecolor{newcolor}{rgb}{.8,.349,.1}

\usepackage{changepage}
\usepackage{framed}
\usepackage{latexsym}
\usepackage{lscape} 
\usepackage{rotating}
\usepackage{tabularx} 
\usepackage[linesnumbered,ruled,vlined,noline,noend]{algorithm2e}
\usepackage{soul} 
\usepackage{balance}
\usepackage{flushend}
\usepackage{microtype}
\usepackage{placeins}

\begin{document}

\begin{frontmatter}

\title{CapsProm: A Capsule Network For Promoter Prediction}

\author[decomufop]{Lauro Moraes\corref{mycorrespondingauthor}}
\ead{lauromoraes@ufop.edu.br}

\author[decomufop]{Pedro Silva}
\author[decomufop]{Eduardo Luz}
\author[decomufop]{Gladston Moreira}

\cortext[mycorrespondingauthor]{Corresponding author}

\address[decomufop]{Computing Department, Federal University of Ouro Preto, Campus Morro do Cruzeiro, Ouro Preto-MG, Brazil}

\begin{abstract}
Locating the promoter region in DNA sequences is of paramount importance in the field of bioinformatics. This is a problem widely studied in the literature, however, not yet fully resolved. Some researchers have presented remarkable results using convolution networks, that allowed the automatic extraction of features from a DNA chain. However, a universal architecture that could generalize to several organisms has not yet been achieved, and thus, requiring researchers to seek new architectures and hyperparameters for each new organism evaluated. In this work, we propose a versatile architecture, based on capsule network, that can accurately identify promoter sequences in raw DNA data from seven different organisms, eukaryotic, and prokaryotic. Our model, the CapsProm, could assist in the transfer of learning between organisms and expand its applicability. Furthermore the CapsProm showed competitive results, overcoming the baseline method in five out of seven of the tested datasets (F1-score). The models and source code are made available at \url{https://github.com/lauromoraes/CapsNet-promoter}.
\end{abstract}


\end{frontmatter}


\section{Introduction}

The central dogma of molecular biology~\citep{crick1970central} describes how DNA acts as a template and provides information to construct functional products, such as proteins molecules. Two main processes are involved in the gene expression task: transcription and translation. Firstly, the transcription process encodes information from a gene, a sequence of DNA, into RNA molecules, or mRNA molecules in eukaryotes, using a similar RNA alphabet. The translation process decodes the information of these RNA molecules into new amino acid sequences of polypeptides.

Around the gene exists various regulatory regions involved in the translation process. These regions may contain DNA elements and motifs that can provide clues to the challenge of predicting where the gene is in the DNA strand. In this sense, understanding the structure of the gene is essential to molecular biology. The location where initiates the transcription of the first DNA nucleotide is called transcription start site (TSS)~\citep{pedersen1999biology}. A relevant region related to the gene transcription process that is located along with upstream of the TSS is the promoter core region. The identification of this essential regulatory region enables the inference of the approximate start position of a gene; thus, many promoter prediction methods and systems have been proposed~\citep{umarov2017recognition, oubounyt2019deepromoter, ohler2002computational, bajic2003dragon, zeng2009towards}.

The promoter core regions may have some common subsequences that indicate where the transcription starts. A subsequence named ``TATA-box'' appears to be conserved in most eukaryotes organisms. It is a motif region with repeated T, and A base pairs (bps) that form a consensus sequence that allowed mismatches. It is related to the TATA-binding protein, found in the pre-initiation process that directs transcriptional starts. Generally, it is located at a position of about 25-35 bps downstream~\citep{hahn1989yeast}. There are other motifs, like the CAAT-box~\citep{stinski1999cytomegalovirus}, which occur at the location 80-110 bps upstream, and the GC-box~\citep{godbey2014introduction}, a transcriptional regulatory element in the position 100 upstream. Besides, another common signal is the CpG islands (CGIs); 300–3000 bps sequences of C and G base pairs that are located within and close to TSS~\citep{deaton2011cpg}.

Methods aiming promoter region identification can approach different types of informations~\citep{singh2015review}. The external information-based methods use a homologous sequence database to align similar sequences and use the matches to determine the region of known genes. The internal information-based methods extract some features from DNA sequences and apply the principle that the properties of the features obtained from promoters are different from the properties of other functional regions~\citep{zeng2009towards, liu2006motif}. Some of these methods focus on specific signals to identify transcription related regions, like the TATA-box~\citep{ohler2002computational}, the CAAT-box~\citep{zeng2009towards}, and the CpG islands~\citep{bajic2003dragon, hannenhalli2001promoter}. 

Promoter sequences are complex and heterogeneous. A limited number of them match the consensus motifs~\citep{gan2012comparison}. Although, the internal information-based methods for promoter prediction can explore other features. Using slides windows is possible to extract distribution statistics from nucleotides in a subsequence~\citep{jabid2007identification, carels2009universal, lin2014ipro54}. Besides, there are the analysis of structural properties features. By the analysis of distinct biophysical functions in the internal interactions of dinucleotides and trinucleotides, it is possible to model nucleotides sequences as correlated structures sequences~\citep{gan2012comparison, carvalho2015impact} or use these biophysical functions to model compositional features~\citep{lin2014ipro54, liu2018ipromoter}. Since the start of the development of new computational methods for the promoter prediction research, several studies applied different  classification methods, like Naive Bayes Approach~\citep{loganantharaj2005recognizing, monteiro2005machine}, K-Nearest Neighbors (KNN)~\citep{gan2012comparison, carvalho2015impact, monteiro2005machine}, Support Vector Machine (SVM)~\citep{gan2012comparison, carvalho2015impact, kasabov2003transductive, monteiro2005machine}, Random Forest (RF)~\citep{carvalho2015impact, liu2018ipromoter, kaladhar2012analysis}, feed-forward Artificial Neural Network (ANN)~\citep{liu2006motif, bajic2002intelligent, loganantharaj2005recognizing, monteiro2005machine, arniker2011dna}. Although deep learning-based techniques achieved outstanding results, outperforming previous state-of-the-art machine learning methods for the promoter prediction problem~\citep{umarov2017recognition}. 

Deep convolution networks, such as those presented in~\citep{umarov2017recognition}, have brought a significant advance to the problem. Therefore, in~\citep{umarov2017recognition}, the authors used one CNN architecture for each organism. The CNN architectures proposed in \citep{umarov2017recognition} did not generalize to different organisms, as also shown here in this work. That is, for each organism, a specific convolution network architecture is required. Here we aim to address this issue by using the CapsNet~\citep{sabour2017dynamic} architecture.

\begin{figure}[!htb]
    \centering
    \includegraphics[width=0.65\linewidth,trim={0 0.9cm 0 0.9cm},clip]{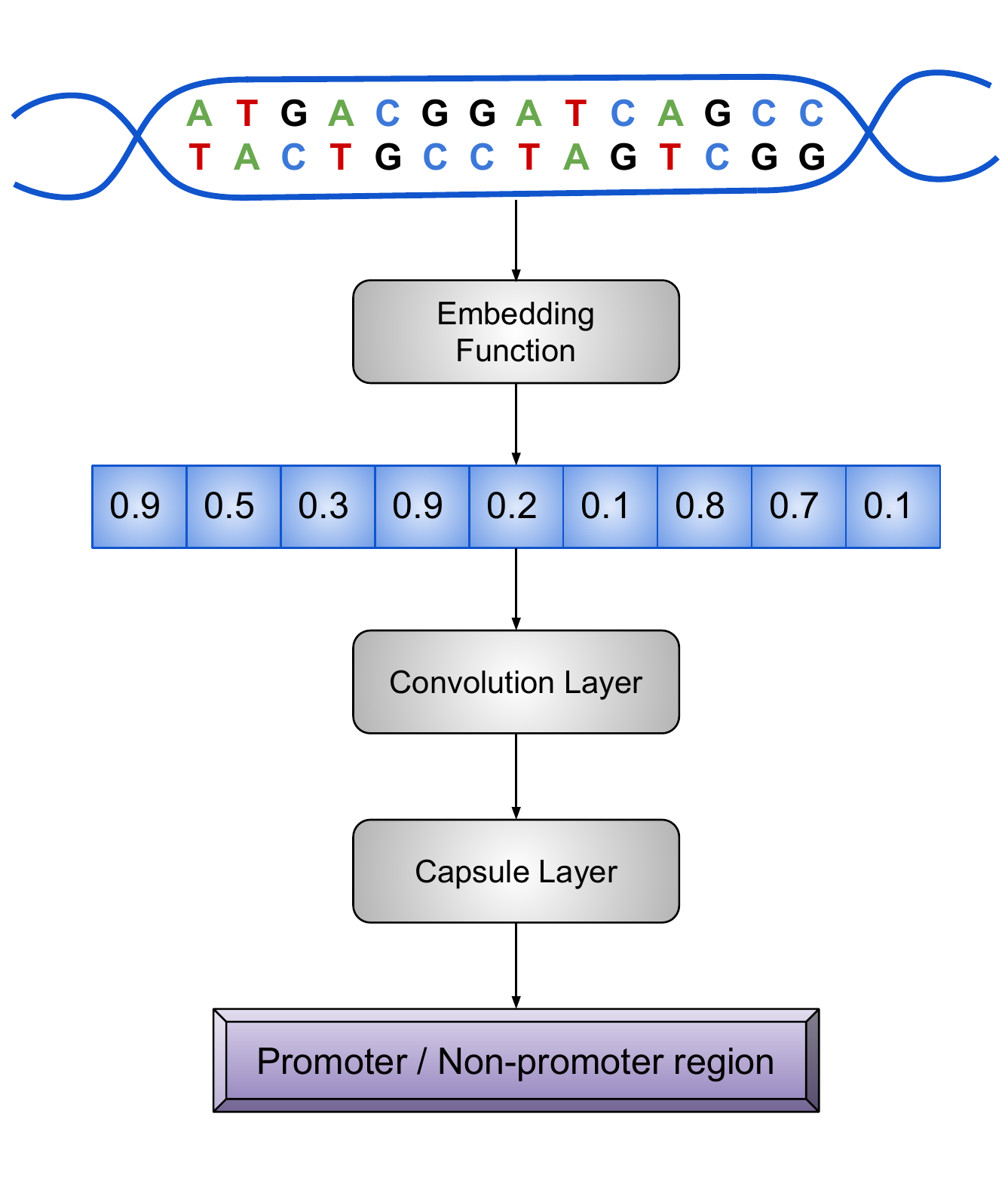}
    \caption{Proposed CapsProm approach.}
    \label{fig:capsprom}
\end{figure}

The CapsNet~\citep{sabour2017dynamic} is a novel approach to represent artificial neurons as capsules, sets of neurons organized as vectors to describe complex interactions between objects hierarchically. Using CapsNet concepts, we present the CapsProm (Figure~\ref{fig:capsprom}), a CapsNet-based architecture to process raw DNA strings sequences and classify these sequences as promoters or non-promoters. CapsProm achieved competitive results when compared to several convoluted network architectures. Therefore, we emphasize here the CapsProm's ability to generalize well to seven different types of organisms. 

In summary, the contributions of this work are:

\begin{itemize}
    \item A competitive CapsNet architecture, the CapsProm, aiming promoter region identification; 
    \item A unique model able to generalize well to seven different organisms;
    \item An open source promoter prediction framework;
    \item A fair and reproducible evaluation protocol;
\end{itemize}

The rest of this work is divided as follow.
In Section~\ref{sec:related}, the background is presented based on the related works.
The methodology is presented in Section~\ref{sec:methods} and the results along with a discussion in Section~\ref{sec:results}.
Finally, a conclusion regarding this work is described in Section~\ref{sec:conclusions}.

\section{Related Works} \label{sec:related}

Convolutional Neural Networks~\citep{lecun1998gradient} (CNN) architectures automatically extract many features from input data through convolutional layers with shared weights, which employ artificial neurons to compute the discrete convolutional function~\citep{lecun1998gradient} over an input field. The Rectified Linear Unit (ReLU) activation function~\citep{nair2010rectified} is used by neurons of CNNs to prevent the saturation of the training process and avoid the vanishing gradients problem~\citep{hochreiter1991untersuchungen}. In order to make the evidence of the most relevant information, a downsample strategy is performed by the pooling layers~\citep{nagi2011max}.

Capsule Neural Network~\citep{sabour2017dynamic} (CapsNet) is an architecture that encodes the properties and the relationships of the features in a hierarchical manner, which shows encouraging results on image classification. The capsule unit concept expands the representational power of a simple neuron unit, encoding information in a multidimensional vector instead of a single scalar. The learning process is made by a function called routing-by-agreement, an iterative process that the lower-level capsules prefer to send the most of activation outputs to higher-level capsules if they agree with the prediction. It is a model that grows in width instead of growing in-depth. More capsules or larger capsules are added to the model rather than add more layers. Unfortunately, it has an expensive computational cost and needs improvements.

Deep learning models have been successfully applied in genomics in several manners~\citep{zou2019primer}.
The DeMo Dashboard~\citep{lanchantin2017deep} toolkit proposed some strategies to visualize and understand learned patterns from deep learning models. Earlier layers from deep learning models can extract patterns that can be related to genomics motifs. Three architectures (a CNN, an RNN, and a CNN-RNN) were used for transcription factor biding (TFBS) site classification task, and the learned parameters were analyzed with the proposed method and compared with know motifs. The results suggest that known and new motifs can be effectively learned through deep learning models.
Recurrent networks were explored in genomics too. A model named KEGRU~\citep{shen2018recurrent} identifies TFBS using a combination of Bidirectional
Gated Recurrent Unit (GRU) network and explores different configurations for k-mer embedding using dense vectors. In the task of predicting RNA secondary structure, an adaptive sequence length based on an LSTM model with an energy-based filter~\citep{lu2019predicting} achieved excellent results.

The CapsNet architecture influenced the development of new capsule-inspired methods for genomics.
iProDNA-CapsNet~\citep{nguyen2019iprodna} presents a capsule network approach to determining the protein-DNA binding
residues, a problem related to the understanding of protein functions and for drug discovery. The authors used the Position-Specific Iterative Basic Local Alignment Search Tool, or simply PSI-BLAST, \citep{altschul1997gapped} to compare 584 non-redundant protein sequences to the Swiss-Prot
database~\citep{bairoch2000swiss} and generated several position-specific scoring matrices (PSSM) used as features to feed the network. Ten CapsNets were trained using a 10-fold stratified data splitting and ensembled together to create the final model. The method demonstrated improvements on several metrics than the state-of-the-art methods.
The CapsNet-SSP~\citep{du2020capsnet} proposed a deep learning framework to identify human saliva-secretory proteins, a non-invasive biomarker. The PSI-BLAST tool converts each input protein sequence to a $1000 x 20$ normalized PSSM. The architecture consists of an ensemble of eight CapsNet with a differently sized 1D convolutional kernel for automatic feature extraction. The reported results outperformed other state-of-the-art deep learning architectures for protein identification application on the tested datasets.
The AC-Caps~\citep{song2020ac} is a hybrid architecture that starts with a joint processing layer formed by an attention mechanism followed by a convolutional layer that fed the capsule layer. The studied problem is predicting RNA-binding protein (RBP) binding sites on the Long non-coding RNA(lncRNA), a region longer than 200 nucleotides that have no protein-encoding function. The lncRNA is essential for many biological processes, and it enables the understanding of the physiological and pathological processes of cancer. The input nucleotides sequences are embedded through a high-order statistical-based encoding method to represent higher dimensional information. The results obtained testing on 31 different datasets showed that the AC-Caps architecture got the best performance over compared methods.

In the literature, we find some works applying deep learning to the promoter classification problem and achieving outstanding results.
The CNNpromoter Data~\citep{umarov2017recognition} is composed of eight datasets (\textit{Escherichia coli} s70, \textit{Bacillus subtilis}, Human TATA, Human non-TATA, Mouse TATA, Mouse non-TATA, \textit{Arabidopsis} TATA, \textit{Arabidopsis} non-TATA) of five distant organisms. A distinct CNN LeNet-like architecture is also proposed in~\citep{umarov2017recognition}, for each dataset, achieving state-of-art results. 
The DeeReCT-PromID~\citep{umarov2019promoter} outperforms previous works for the exact identification of TSS in long human genomic sequences using a CNN and a novel method that creates a challenging negative dataset through an interactive selection of previous false positives examples. 
DeePromoter~\citep{oubounyt2019deepromoter} is a promoter prediction tool that combines a CNN and an LSTM to process short eukaryotic sequences for the classification of human and mouse promoter sequences. The authors proposed a method to construct a new synthetic negative sample applying some random modifications to a positive sample copy. This technique enabled the creation of a new negative dataset that has a more similar sequence logo~\citep{schneider1990sequence} to the positive dataset.

\section{Proposed approach}
\label{sec:methods}

In this section, it is presented the proposed approach, the datasets and the metrics used to report the results.

\subsection{Datasets}

All seven datasets used in this work are acquired from the supplementary materials of the CNNPromoter project\footnote{\url{https://github.com/solovictor/CNNPromoterData}} in GitHub. The \textit{Bacillus subtilis} (Bacillus), and the \textit{Escherichia coli s}70 (Ecoli) datasets are from two different bacterias, that is, prokaryotic organisms. The remaining datasets are from eukaryotic organisms. The \textit{Arabidopsis}\_tata and \textit{Arabidopsis}\_non\_tata datasets contains sequences from a plant specie, and Mouse\_tata, Mouse\_non\_tata, and Human\_non\_tata datasets are from mammals species. The datasets present sequences with a length of $81bp$ for prokaryotic organisms and $251bp$ for eukaryotic. Consider the benchmark dataset $\mathbb{D} = \mathbb{D}^{+} \cup  \mathbb{D}^{-}$, where $\mathbb{D}^{+}$ is the positive subset with all promoters sequences samples and the negative subset $\mathbb{D}^{-}$ contains all non-promoters examples. 

Table~\ref{tab:dbsize} summarizes all the details about each dataset. 
It is possible to observe that all datasets are imbalanced, with a greater number of non-promoter sequences.

\subsection{DNA Sequence Representation}

Adenine (A), cytosine (C), guanine (G), and thymine (T) are the four nucleotides that compound the alphabet $V=\{A, C, G, T\}$ which is the base for the deoxyribonucleic acid (DNA). 
A DNA sample compounded by $l_{S} \in \mathbb{N}^{*}$ base pairs (bp) is denoted by $S \in V^{l_{S}}$ and may be represented as a sequence $S = s_{1} ... s_{i} ... s_{l_{S}}$, where $s_{i}$ is the \textit{i}-th nucleotide on sequence $S$.

The neural networks only accept numerical values as inputs. Categorical data, such as nucleotides sequences, must be appropriately converted before feeding the network. Two principal representations can be used: one-hot-encoding vectors and embeddings.

The one-hot-encoding strategy is widely used to represent nucleotides sequences as sparse binary vectors for deep learning approaches~\citep{umarov2017recognition, nguyen2019iprodna, levy2020investigation}. 
To represent a sample, let be $O$ a function that converts a single nucleotide character in a one-hot 4-dimensional vector. So, the representation of each nucleotide is: $O(A)=<1, 0, 0, 0>$, $O(C)=<0, 1, 0, 0>$, $O(G)=<0, 0, 1, 0>$ and $O(T)=<0, 0, 0, 1>$.

An embedding is a learned dense fixed size low-dimen-sional continuous vector that maps a discrete variable into a continuous one. Consider a vector length $l_{\epsilon} \in \mathbb{N}^{*}$, and let be $\epsilon$ an embed function to learn and convert a nucleotide to a $l_{\epsilon}$-dimensional continuous vector. In this manner, $\epsilon(s)=<e_{1}^{s}, ..., e_{i_{\epsilon}}^{s}, ..., e_{l_{\epsilon}}^{s}>$, in which $e_{i_{\epsilon}}^{s} \in \mathbb{R}$ is the $i_{\epsilon}$-th continuous learned value for the nucleotide $s \in V$ in the embbeded vector. The $e_{i_{\epsilon}}^{s}$ value is learned during the training process, like other network parameters. Due to the great advance that embedding techniques have brought to Natural Language Processing (NLP) problems, in this work, we propose the use of embeddings to encode the nucleotides sequences.

\begin{table}[!htb]
    \caption{Datasets characteristics.}
    \label{tab:dbsize}
    \centering
        \begin{tabular}{|l|c|c|c|c|}
            \hline
            Dataset & $|\mathbb{S}^{-}|$ & $|\mathbb{S}^{+}|$ & bp & Type \\ \hline \hline
            \textit{Arabidopsis} non-TATA   & 11459 & 5905 & 251 & Eukaryotic \\ \hline
            \textit{Arabidopsis} TATA       & 2879 & 1497 & 251 & Eukaryotic \\ \hline
            \textit{Bacillus subtilis}      & 1000 & 373 & 81 & Prokaryotic \\ \hline
            \textit{Escherichia coli s}70   & 3000 & 839 & 81 & Prokaryotic \\ \hline
            Human non-TATA                  & 27731 & 19811 & 251 & Eukaryotic \\ \hline
            Mouse non-TATA                  & 24822 & 16283 & 251 & Eukaryotic \\ \hline
            Mouse TATA                      & 3530 & 1255 & 251 & Eukaryotic \\ \hline
        \end{tabular}
\end{table}

\subsection{Capsule networks}

The Capsule Network (CapsNet) \citep{sabour2017dynamic} deconstruct the input in a hierarchical manner, preserving the relationships between identified entities.
CapsNet present an approach that group neurons into a vector structure called capsule, expanding the representational capabilities by performing a non-linear map from vector to vector. 

The neurons' activity in a capsule act as the instantiation parameters to represent a specific entity. 
The length or magnitude of the activation vector is associated with the probability of that entity's existence.

The CapsNet does not use the pooling downsample strategy. Instead, the capsules capture the equivariance of the objects. This property can reduce the necessity for a large number of examples to be provided to the network, and the use of techniques like data augmentation ~\citep{sabour2017dynamic}.

The proposed CapsNet architecture is called CapsProm and be seen in Figure~\ref{fig:capsprom}. In this new architecture, the input nucleotide sequence is provided to an embedding layer that learns and codifies the vectors to represent the sequence's nucleotides. Further, a standard convolutional layer receives the embedded vector representations and extracts the low-level local features from the encoded DNA input. It consists of $256$ filters with size $1 \times 9$, stride of $1$ and Rectified Linear Units (ReLU)~\citep{nair2010rectified} as activation function. After this point, initiates a group of stages that characterizes the PrimaryCaps (PC) of the CapsNet.

On PrimaryCaps, a convolutional layer, with $256$ filter with dimensions of $1 \times 9$ and stride $2$, extracts high-level features, and then, the resulting volume is reshaped into 32 channels of 8-dimensional capsules. Each of these capsules of the PC is constituted of eight instantiations parameters and is represented by a vector in 8-dimensional space. A squash function is applied over these vectors, normalizing the lengths in the range of $[0, 1]$ without modifying the relative proportions between them. After the squash, a new reshape is applied, transforming the volume into a vector of capsules to be passed to the next group of stages, the DigitsCaps (DC).

On DigitsCaps, each capsule of the PrimaryCaps is associated with a weight matrix that expands its dimensions from 8 to 16. Each 16-dimensional vector is associated with a coupling coefficient for each capsule of the DigitsCaps. A weighted sum is performed between these coefficients and the new 16-dimensional capsules, and then a squash function normalizes the magnitudes, generating the resulting capsules of the DC. The DigitsCaps has a total of capsules equal to the number of different classes that the problem has, so in this work, the DC has two capsules (promoter and non-promoter sequence).

To adjust the coupling coefficients values, an approach proposed in~\citep{sabour2017dynamic}, called routing by agreement algorithm, is considered. It verifies the conformity between the capsules in the PC and those related in the DC. The coefficients that link capsules with an agreement in the orientation are reinforced and those that disagree are lowered.

\begin{figure}[!htb]
    \centering
    \includegraphics[width=0.98\linewidth]{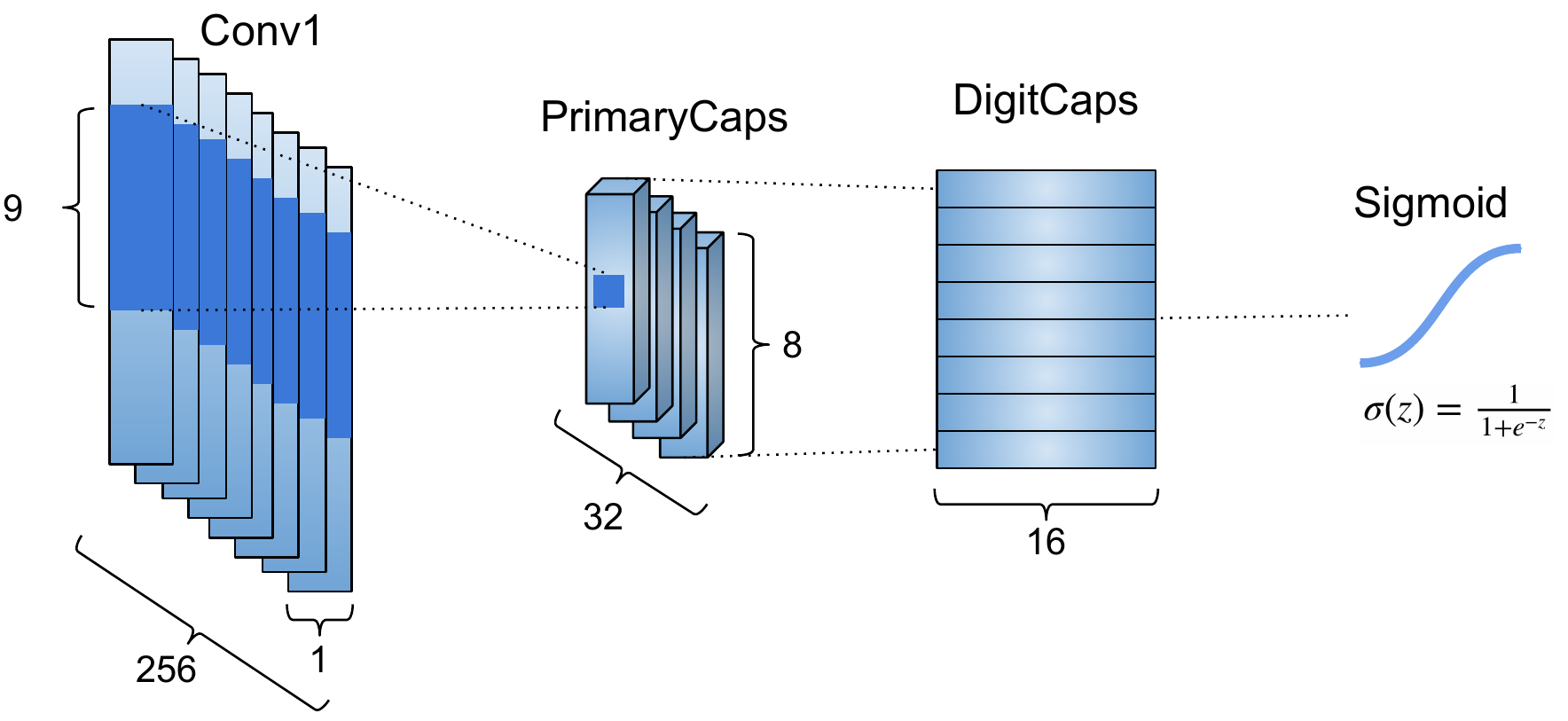}
    \caption{Proposed CapsProm architecture.}
    \label{fig:capspromscheme}
\end{figure}

The capsules on the DigitsCaps are linked to a fully connected layer with $128$ neurons with ReLU activation function, and then to a single neuron with a sigmoid activation function to output the network value. The margin loss function calculates the loss of this network~\cite{sabour2017dynamic}. Table~\ref{tab:capsprom} summarizes all the stages of the proposed CapsProm and Figure~\ref{fig:capspromscheme} shows it graphically.

\begin{table}[!htb]
    \caption{Definition of the CapsProm architecture.}
    \label{tab:capsprom}
    \centering
        \begin{tabular}{|l|c|}
        \hline
        \textbf{Stage} & \textbf{Definition}  \\ \hline \hline
        Input & Dimensions: ($1 \times 251$) or ($1 \times 81$) \\ \hline
        Embedding & Dense vector dimensions: 9 \\ \hline
        Convolution & 256 filters ($1 \times 9$); Stride=1; ReLu \\ \hline
        PC - Convolution & 256 filters ($1 \times 9$); Stride=2; ReLu \\ \hline
        PC - Reshape & 8-dimensional vectors in 32 channels \\ \hline
        PC - Squash & Squashing of 8-dimensional vectors \\ \hline
        DC - Weights & Multiplication by the weight matrix \footnotemark \\ \hline
        DC - Coupling and sum & Weighted sum in each DC capsule \\ \hline
        DC - Squash & Squashing of 16-dimensional vectors \\ \hline
        Fully Connected & 1 neuron; Sigmoid \\ \hline
        \end{tabular}
\end{table}

\footnotetext{Transforms 8D vectors to 16D.}

\subsection{Convolutional Neural Network}
\begin{figure}[!htb]
    \centering
    \begin{tabular}{c}
        \includegraphics[page=4,width=0.8\linewidth,trim={0 0.5cm 0 0.5cm},clip]{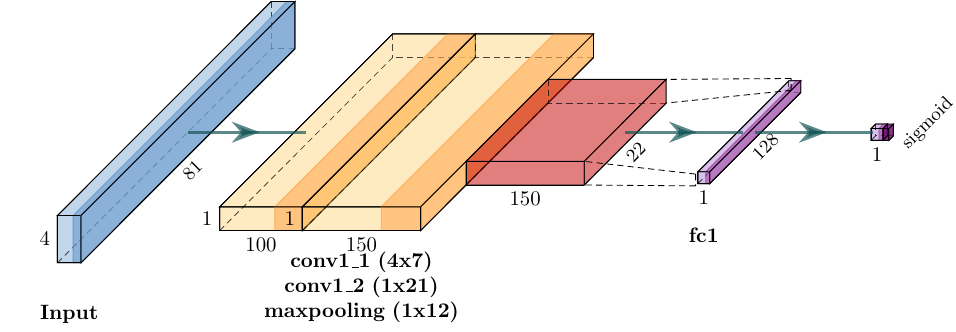} \\
        (d) \\
        \includegraphics[page=5,width=0.8\linewidth,trim={0 0.5cm 0 0.5cm},clip]{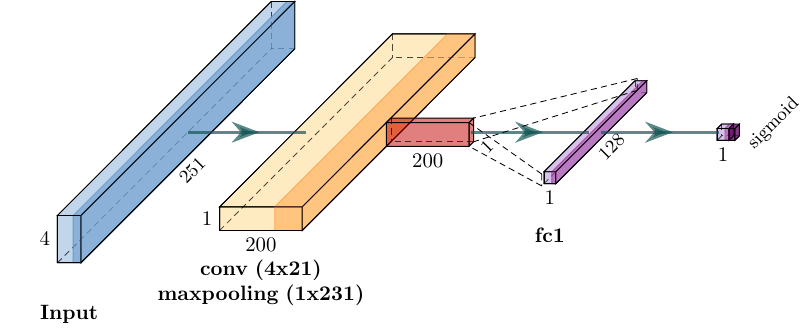} \\
        (e) \\
        \includegraphics[page=6,width=0.8\linewidth,trim={0 0.5cm 0 0.5cm},clip]{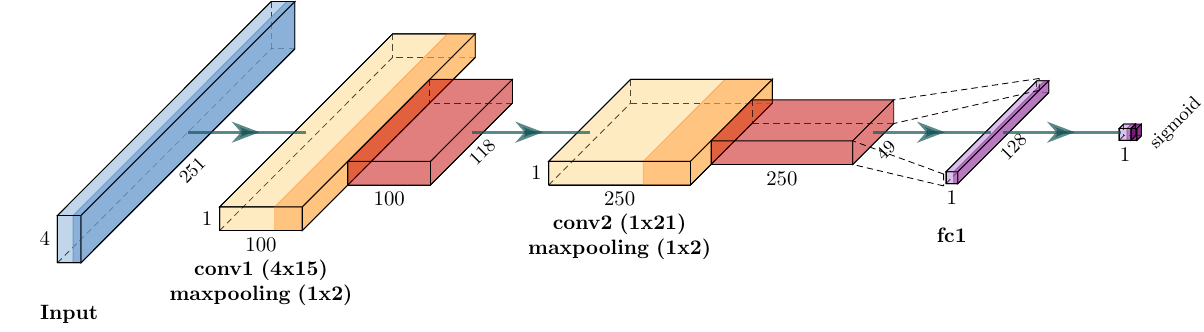} \\
        (f)\\
        \includegraphics[page=7,width=0.8\linewidth,trim={0 0.5cm 0 0.5cm},clip]{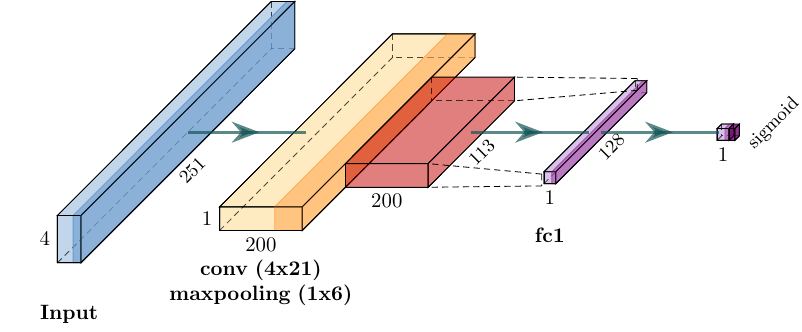} \\
        (g) 
    \end{tabular}
    \caption{Proposed convolutions architectures implemented to process: 
        (a) \textit{Arabidopsis}\_non\_tata dataset,
        (b) \textit{Arabidopsis}\_tata dataset,
        (c) \textit{Bacillus subtilis} dataset,
        (d) \textit{Escherichia coli s}70 dataset,
        (e) \textit{Human}\_non\_tata dataset,
        (f) \textit{Mouse}\_non\_tata dataset,
        (g) \textit{Mouse}\_tata dataset. Convolution stride is equal to one and in MaxPooling is two.}
    \label{fig:cnns}
\end{figure}

In this work, we implemented the Convolutional Neural Networks (CNN) models based on the literature~\citep{umarov2017recognition}. We use CNNProm to identify the re-implementation of the CNNs. These implemented models were used as baseline to our proposed capsule-network-based model. All CNN models are feed by a sequence of one-hot-encoded nucleotides. Following the guides in the literature, we implemented a unique CNN model for each tested dataset. We present all the specifics CNNProms configurations in Figure~\ref{fig:cnns}.

\begin{figure}[!htb]
    \centering
    \begin{tabular}{c}
        \includegraphics[page=1,width=0.85\linewidth,trim={0 0.5cm 0 0.5cm},clip]{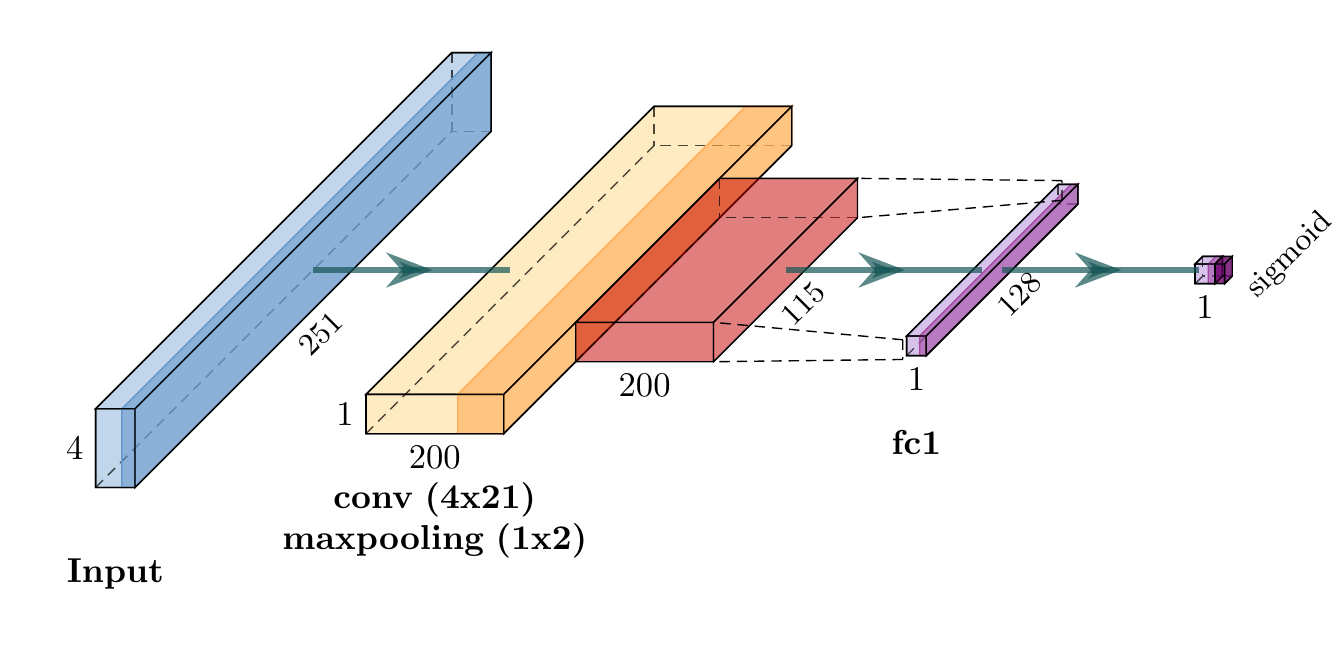} \\
        (a)\\
        \includegraphics[page=2,width=0.85\linewidth,trim={0 0.5cm 0 0.5cm},clip]{caps-nets.pdf} \\
        (b)\\
        \includegraphics[page=3,width=0.85\linewidth,trim={0 0.5cm 0 0.5cm},clip]{caps-nets.pdf} \\
        (c) \\
    \end{tabular}
\end{figure}

\FloatBarrier

\subsection{K-Fold Cross-Validation evaluation}

As shown in Table~\ref{tab:dbsize}, neither dataset utilized in this work has a large amount of available data, considering the machine learning context. A fair manner to evaluate the performance of an approach with unseen data by the models with limited and unbalanced data samples is to use the cross-validation procedure. 
Firstly, the dataset samples are shuffled. Defining a parameter $k \in \mathbb{N}^{*}$, the original dataset is split into $k$ disjoint stratified partitions of approximately equal size. 
In a process of $k$ iterations, the k$^{th}$ partition is used for testing and all the remaining are used for training.
In the experiments with CNNs and the CapsProm, a $5$-fold cross-validation was used to evaluate the proposed approaches.

\subsection{Evaluation Measures}
\label{subsec:metrics}

After the evaluation scenario is described, the metrics used to report the results must be defined.

The Promoter Classification problem is a binary classification problem: 1 (one) for a predicted promoter sequence, 0 (zero) otherwise. The classification has four types of results: a true positive ($TP$) result, where the model predicts as promoter a promoter sequence; a true negative ($TN$), where the model predicts as non-promoter a non-promoter sequence; a false positive ($FP$), where the model predicts as promoter a non-promoter sequence; and a false negative ($FN$), where the model predicts as non-promoter a promoter sequence.

Based on the four types of results, five metrics are adopted.
The precision
\begin{equation}\label{metric:prec}
    Prec = \frac{TP}{TP+FP}
\end{equation}

\noindent measures the relevance of the predicted true positives.
The recall or sensibility
\begin{equation}\label{metric:sn}
    Sn = \frac{TP}{TP+FN}
\end{equation}

\noindent measures the proportion of the true positives.
The F1-score metric 
\begin{equation}\label{metric:f1}
    F1 = 2 \times \frac{Prec \times Sn}{Prec + Sn}
\end{equation}

\noindent is a harmonic mean of the precision and recall, or sensibility. 
The specificity
\begin{equation}\label{metric:sp}
    Sp = \frac{TN}{TN+FP}
\end{equation}

\noindent measures the proportion of the true negatives.
The accuracy 
\begin{equation}\label{metric:acc}
    Acc= \frac{TP+TN}{TP+TN+FP+FN}
\end{equation}

\noindent reflects the proportion of the correct predictions. 
Finally, the Matthews correlation coefficient metric (also known as Pearson's phi coefficient)
\begin{equation}\label{metric:mcc}
    Mcc = \frac{TP \times TN + FP \times FN}{\sqrt[]{(TP+FP)(TP+FN)(TN+FP)(TN+FN)}}
\end{equation}

\noindent  considers all four classification results types and provides a balanced measure, even with a non-balanced amount of data by class in the test dataset.

The image of the metric function $Mcc$ is in the range $[-1, 1]$, where $1$ means a perfect prediction, $0$ indicates random prediction, and $-1$ designates a total disparity between the predictions and the true classes labels. All other metrics functions have images on the range $[0, 1]$, where $0$ indicates a worse prediction performance of the evaluated model and $1$ shows a good generalization of the model within the test dataset.

\section{Results and Discussion}
\label{sec:results}

In this section, we compare the results among the CNN models (CNNProm) and the proposed CapsProm.

\subsection{CNNProms analysis}
\label{subsec:comp_cnn}
Before comparing the capsule approach with the convolutional approach, we evaluated our CNN implementation (CNNProm) that follows the models proposed in the CNNPromoter project. Considering ``M1'' as the reported mean results of the CNNPromoter project and ``M2'' as mean results of our implementation (CNNProm), Table~\ref{tab:results_metrics} compares $Sn$, $Sp$ and $Mcc$ metrics of these two models.

\begin{table}[!htb]
    \caption[Mean values of metrics obtained from 5-fold cross-validation on each dataset. ``M1'' shows results from the paper of CNNPromoter project and ``M2'' presents obtained results for our implementation of CNN model (CNNProm) based on CNNPromoter project specifications.]{Mean values of metrics obtained from 5-fold cross-validation on each dataset. ``M1'' shows results from the paper of CNNPromoter project and ``M2'' presents obtained results for our implementation of CNN model (CNNProm) based on CNNPromoter project specifications.}
    \centering
        \begin{tabular}{|l|c|c|c|c|c|c|}
            \hline
            Dataset 			& \multicolumn{2}{|c|}{Sn}   & \multicolumn{2}{|c|}{Sp} & \multicolumn{2}{|c|}{Mcc}  \\ \hline \hline
            
             & M1 & M2 & M1 & M2 & M1 & M2      \\ \hline \hline
            
            \textit{Arabidopsis}\_non\_tata	& \textbf{0,94} & 0,91 & 0,94 & \textbf{0,96} & 0,86 & \textbf{0,88}      \\ \hline
            \textit{Arabidopsis}\_tata	& 0,95 & \textbf{0,97} & 0,97 & \textbf{0,98} & 0,91 & \textbf{0,95}      \\ \hline
            \textit{Bacillus subtilis}	& \textbf{0,91} & 0,88 & 0,95 & 0,95 & \textbf{0,86} & 0,83      \\ \hline
            \textit{Escherichia coli s}70	& \textbf{0,90} & 0,87 & 0,96 & \textbf{0,97} & 0,84 & \textbf{0,86}      \\ \hline
            Human\_non\_tata	& 0,90 & 0,90 & 0,98 & 0,98 & 0,89 & \textbf{0,90}      \\ \hline
            Mouse\_non\_tata	& 0,88 & 0,88 & 0,94 & \textbf{0,95} & 0,83 & \textbf{0,84}      \\ \hline
            Mouse\_tata	& \textbf{0,97} & 0,94 & 0,97 & \textbf{0,98} & \textbf{0,93} & 0,92      \\ \hline
        \end{tabular}
    \label{tab:results_metrics}
\end{table}

As one can see, our re-implemented convolutional models achieved close results to the ones reported in (\cite{umarov2017recognition}. Our implementation achieved better figures on five out of seven evaluation scenarios. Thus, we consider that our implementation of the model inspired on the CNNPromoter is a fair candidate to be compared with our implementation of the CapsNet approach, the CapsProm. The CNNPromoter souce code is available at \url{https://github.com/lauromoraes/CapsNet-promoter}.

\subsection{CNNProms vs. CapsProms analysis}
\label{subsec:comp_conv_caps}

All metrics were obtained from 5-fold cross-validation. The same partitions subsets were used for both CNNProm and CapsProm training/test iterations. Using the boxplots, it is possible to observe the variation of both methods' predictions performance on the tested partitions. Results for all six metrics presented in Subsection~\ref{subsec:metrics} are presented in the graphs.

\begin{figure}[!htb]
    \centering
    \includegraphics[width=0.9\linewidth]{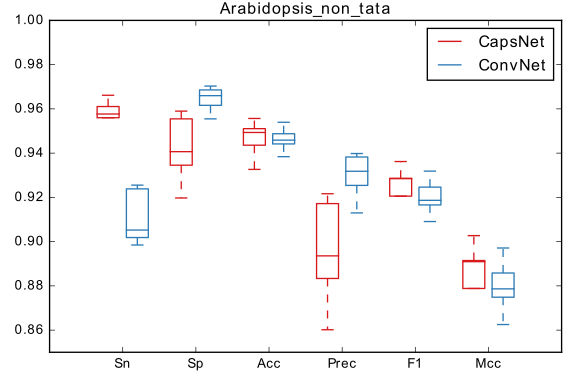}
    \caption{Metrics obtained for CapsProm (red boxes) and for CNNProm (blue boxes) in 5-fold cross-validation on \textit{Arabidopsis}\_non\_tata dataset.}\label{fig:boxplot_arabidopsis_non_tata}
\end{figure}

Figure~\ref{fig:boxplot_arabidopsis_non_tata} presents metrics on the \textit{Arabidopsis}\_non\_tata dataset. As one can see, the results for metrics $Sn$, $Sp$ and $Prec$ shows an apparently significant difference. The $Mcc$ results are slightly higher for the CapsNet model.

\begin{figure}[!ht]
    \centering
    \includegraphics[width=0.9\linewidth]{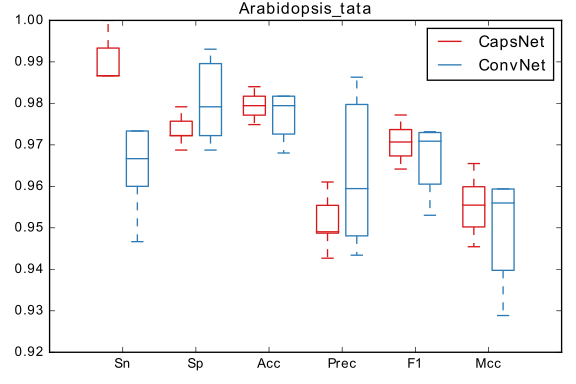}
    \caption{Metrics obtained for CapsProm (red boxes) and for CNNProm (blue boxes) in 5-fold cross-validation on \textit{Arabidopsis}\_tata dataset.}\label{fig:boxplot_arabidopsis_tata}
\end{figure}

The metrics calculated on the \textit{Arabidopsis}\_tata database are shown in Figure \ref{fig:boxplot_arabidopsis_tata}. The averages for the $Mcc$ metrics are very close. However, analyzing the CNNProm model's variance shows lower boundaries than those of CapsProm.

\begin{figure}[!ht]
    \centering
    \includegraphics[width=0.9\linewidth]{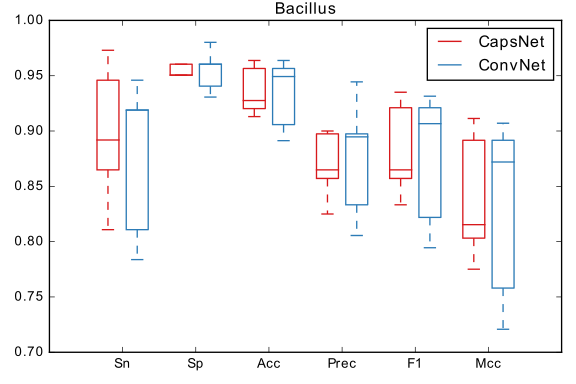}
    \caption{Metrics obtained for CapsProm (red boxes) and for CNNProm (blue boxes) in 5-fold cross-validation on \textit{Bacillus subtilis} dataset.}\label{fig:boxplot_bacillus}
\end{figure}

Regarding the metrics shown in Figure \ref{fig:boxplot_bacillus}, the CapsProm model is less susceptible to variance em relation to the CNNProm. However, the CNNProm model averages show superiority.

\begin{figure}[!ht]
    \centering
    \includegraphics[width=0.9\linewidth]{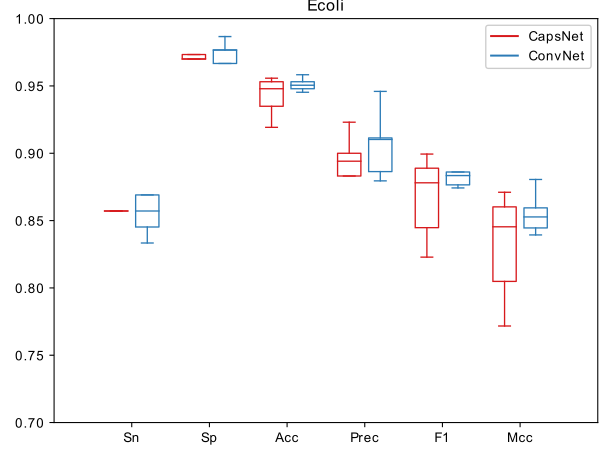}
    \caption{Metrics obtained for CapsProm (red boxes) and for CNNProm (blue boxes) in 5-fold cross-validation on \textit{Escherichia coli s}70 dataset.}\label{fig:boxplot_ecoli}
\end{figure}

The results shown in Figure \ref{fig:boxplot_ecoli} indicates larger variance in relation to the $F1$ and $Mcc$ metrics of the CapsProm model. However, their averages are close to the CNNProm model.

\begin{figure}[!ht]
    \centering
    \includegraphics[width=0.9\linewidth]{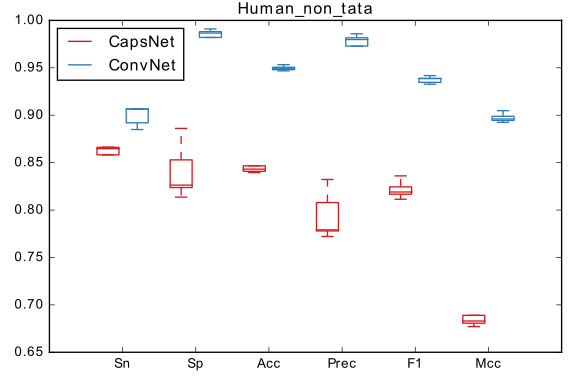}
    \caption{Metrics obtained for CapsProm (red boxes) and for CNNProm (blue boxes) in 5-fold cross-validation on  dataset.}\label{fig:boxplot_human}
\end{figure}

Figure \ref{fig:boxplot_human} shows the values of the metrics calculated for the Human\_non\_tata database.
In this comparison, the difference between the CapsProm and CNNProm models is very evident.
The CapsProm did not performed well in this evaluation.
Perhaps this low performance is related to the high rate of false positives generated in the CapsProm model's predictions for this database.

Concerning all the tests carried out in this work, it was in the database \textit{Human\_non\_tata} that the biggest difference occurred between the metrics evaluated of the tested models.

\begin{figure}[!ht]
    \centering
    \includegraphics[width=0.9\linewidth]{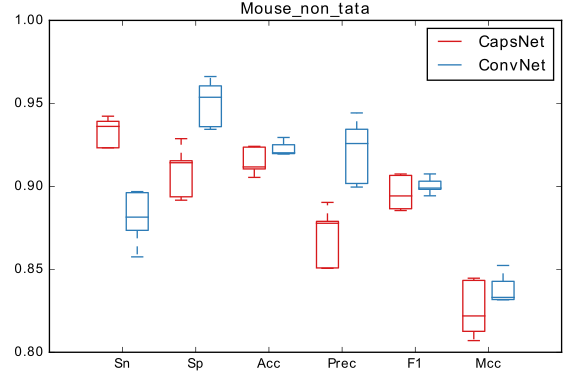}
    \caption{Metrics obtained for CapsProm (red boxes) and for CNNProm (blue boxes) in 5-fold cross-validation on  dataset.}\label{fig:boxplot_mouse_non_tata}
\end{figure}

The metrics $Sn$, $Sp$, and $Prec$ calculated on the dataset Mouse\_non\_tata database are quite different between the models evaluated. The $Mcc$ metric values for both are close, although the CNNProm model has a higher average and a smaller variance. Figure \ref{fig:boxplot_mouse_non_tata} shows the results obtained in the evaluation of this database.

\begin{figure}[!ht]
    \centering
    \includegraphics[width=0.9\linewidth]{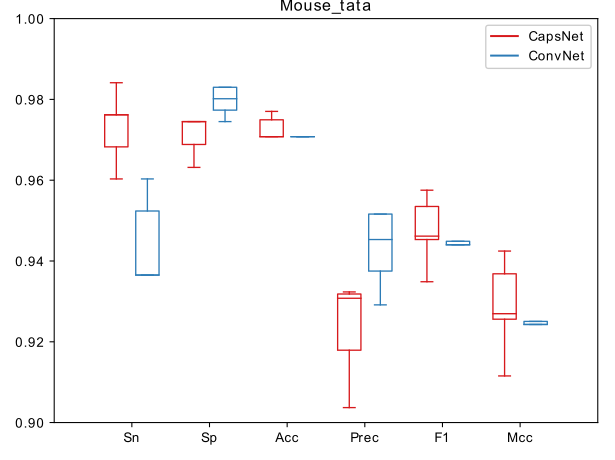}
    \caption{Metrics obtained for CapsProm (red boxes) and for CNNProm (blue boxes) in 5-fold cross-validation on  dataset.}\label{fig:boxplot_mouse_tata}
\end{figure}

The evaluation of the Mouse\_tata database is shown in Figure \ref{fig:boxplot_mouse_tata}. The averages of the $F1$ and $Mcc$ metrics of the CapsProm model are slightly higher than those of the CNNProm model, although their variances are greater.

This work results suggest that the capsule and convolutional architecture have very close predictive capabilities, with some databases standing out slightly more than the other. In just one database, the Human\_non\_tata, the convolutional model far surpassed the capsule model. In general, both models achieved outstanding results in their tests.

\section{Conclusions} \label{sec:conclusions}

The capsules networks~\citep{sabour2017dynamic} has not yet been extensively evaluated in the current genomics literature. In this work, we presented an adaptation of the CapsNet architecture to classify promoter regions using raw DNA strands, called here as CapsProm. As far as the authors knowledge, this work presents the first application of the capsules approach in this scenario. 
We also re-implemented state-of-the-art CNN models, based on the CNNPromoter project~\citep{umarov2017recognition}, for literature comparison purposes. These CNN models were compared to our CapsProm to predict promoter sequences in seven datasets of distinct organism types. CapsProm achieved competitive performance and ouvercomes the CNNProm in five out of seven datasets (in terms of F1-socre). However, CapsProm proved to be superior in one important aspect, which is greater generalization power. With just a single architecture, CapsProm achieved competitive results for all organisms. We believe that this greater generalization power can favor the transfer of learning to other organisms and facilitate future use in the laboratory.

\section*{Acknowledgment}

The authors would like to thank UFOP and funding Brazilian agencies CAPES, FAPE-MIG, and CNPq. We want to express our sincere gratitude for the collaboration of the Multi-user Bioinformatics Laboratory of the Nucleus of Research in Biological Sciences, UFOP. We gratefully acknowledge the support of NVIDIA Corporation with the donation of the Titan X Pascal GPU used for this research.


\bibliography{main}

\begin{thebibliography}{10}
\expandafter\ifx\csname url\endcsname\relax
  \def\url#1{\texttt{#1}}\fi
\expandafter\ifx\csname urlprefix\endcsname\relax\def\urlprefix{URL }\fi
\expandafter\ifx\csname href\endcsname\relax
  \def\href#1#2{#2} \def\path#1{#1}\fi

\bibitem{crick1970central}
F.~Crick, Central dogma of molecular biology, Nature 227~(5258) (1970)
  561--563.

\bibitem{pedersen1999biology}
A.~G. Pedersen, P.~Baldi, Y.~Chauvin, S.~Brunak, The biology of eukaryotic
  promoter prediction—a review, Computers \& chemistry 23~(3-4) (1999)
  191--207.

\bibitem{umarov2017recognition}
R.~K. Umarov, V.~V. Solovyev, Recognition of prokaryotic and eukaryotic
  promoters using convolutional deep learning neural networks, PloS one 12~(2)
  (2017) e0171410.

\bibitem{oubounyt2019deepromoter}
M.~Oubounyt, Z.~Louadi, H.~Tayara, K.~T. Chong, Deepromoter: Robust promoter
  predictor using deep learning, Frontiers in genetics 10.

\bibitem{ohler2002computational}
U.~Ohler, G.-c. Liao, H.~Niemann, G.~M. Rubin, Computational analysis of core
  promoters in the drosophila genome, Genome biology 3~(12) (2002)
  research0087--1.

\bibitem{bajic2003dragon}
V.~B. Bajic, S.~H. Seah, Dragon gene start finder: an advanced system for
  finding approximate locations of the start of gene transcriptional units,
  Genome research 13~(8) (2003) 1923--1929.

\bibitem{zeng2009towards}
J.~Zeng, S.~Zhu, H.~Yan, Towards accurate human promoter recognition: a review
  of currently used sequence features and classification methods, Briefings in
  bioinformatics 10~(5) (2009) 498--508.

\bibitem{hahn1989yeast}
S.~Hahn, S.~Buratowski, P.~A. Sharp, L.~Guarente, Yeast tata-binding protein
  tfiid binds to tata elements with both consensus and nonconsensus dna
  sequences, Proceedings of the National Academy of Sciences 86~(15) (1989)
  5718--5722.

\bibitem{stinski1999cytomegalovirus}
M.~F. Stinski, Cytomegalovirus promoter for expression in mammalian cells, Gene
  expression systems (1999) 211--233.

\bibitem{godbey2014introduction}
W.~T. Godbey, An Introduction to Biotechnology: The Science, Technology and
  Medical Applications, Elsevier, 2014.

\bibitem{deaton2011cpg}
A.~M. Deaton, A.~Bird, Cpg islands and the regulation of transcription, Genes
  \& development 25~(10) (2011) 1010--1022.

\bibitem{singh2015review}
S.~Singh, S.~Kaur, N.~Goel, A review of computational intelligence methods for
  eukaryotic promoter prediction, Nucleosides, Nucleotides and Nucleic Acids
  34~(7) (2015) 449--462.

\bibitem{liu2006motif}
D.~Liu, X.~Xiong, B.~DasGupta, H.~Zhang, Motif discoveries in unaligned
  molecular sequences using self-organizing neural networks, IEEE Trans. Neural
  Networks 17~(4) (2006) 919--928.

\bibitem{hannenhalli2001promoter}
S.~Hannenhalli, S.~Levy, Promoter prediction in the human genome,
  Bioinformatics 17~(suppl\_1) (2001) S90--S96.

\bibitem{gan2012comparison}
Y.~Gan, J.~Guan, S.~Zhou, A comparison study on feature selection of dna
  structural properties for promoter prediction, BMC bioinformatics 13~(1)
  (2012) 4.

\bibitem{jabid2007identification}
T.~Jabid, F.~Anwar, S.~M. Baker, M.~Shoyaib, Identification of promoter through
  stochastic approach, in: 2007 10th international conference on computer and
  information technology, IEEE, 2007, pp. 1--4.

\bibitem{carels2009universal}
N.~Carels, R.~Vidal, D.~Fr{\'\i}as, Universal features for the classification
  of coding and non-coding dna sequences, Bioinformatics and biology insights 3
  (2009) BBI--S2236.

\bibitem{lin2014ipro54}
H.~Lin, E.-Z. Deng, H.~Ding, W.~Chen, K.-C. Chou, ipro54-pseknc: a
  sequence-based predictor for identifying sigma-54 promoters in prokaryote
  with pseudo k-tuple nucleotide composition, Nucleic acids research 42~(21)
  (2014) 12961--12972.

\bibitem{carvalho2015impact}
S.~G. Carvalho, R.~Guerra-S{\'a}, L.~H. de~C~Merschmann, The impact of sequence
  length and number of sequences on promoter prediction performance, BMC
  bioinformatics 16~(S19) (2015) S5.

\bibitem{liu2018ipromoter}
B.~Liu, F.~Yang, D.-S. Huang, K.-C. Chou, ipromoter-2l: a two-layer predictor
  for identifying promoters and their types by multi-window-based pseknc,
  Bioinformatics 34~(1) (2018) 33--40.

\bibitem{loganantharaj2005recognizing}
R.~Loganantharaj, Recognizing transcription start site (tss) of plant
  promoters, in: International Conference on Information Technology: Coding and
  Computing (ITCC'05)-Volume II, Vol.~1, IEEE, 2005, pp. 20--25.

\bibitem{monteiro2005machine}
M.~I. Monteiro, M.~C. de~Souto, L.~M. Gon{\c{c}}alves, L.~F. Agnez-Lima,
  Machine learning techniques for predicting bacillus subtilis promoters, in:
  Brazilian Symposium on Bioinformatics, Springer, 2005, pp. 77--84.

\bibitem{kasabov2003transductive}
N.~Kasabov, S.~Pang, Transductive support vector machines and applications in
  bioinformatics for promoter recognition, in: International Conference on
  Neural Networks and Signal Processing, 2003. Proceedings of the 2003, Vol.~1,
  IEEE, 2003, pp. 1--6.

\bibitem{kaladhar2012analysis}
D.~Kaladhar, T.~U. Devi, P.~Lakshmi, R.~H. Reddy, P.~N. Rao, et~al., Analysis
  of e. coli promoter regions using classification, association and clustering
  algorithms, in: Proceedings of the International Conference on Information
  Systems Design and Intelligent Applications 2012 (INDIA 2012) held in
  Visakhapatnam, India, January 2012, Springer, 2012, pp. 169--177.

\bibitem{bajic2002intelligent}
V.~B. Bajic, A.~Chong, S.~H. Seah, V.~Brusic, An intelligent system for
  vertebrate promoter recognition, IEEE Intelligent Systems 17~(4) (2002)
  64--70.

\bibitem{arniker2011dna}
S.~B. Arniker, H.~K. Kwan, N.-F. Law, D.~P.-K. Lun, Dna numerical
  representation and neural network based human promoter prediction system, in:
  2011 Annual IEEE India Conference, IEEE, 2011, pp. 1--4.

\bibitem{sabour2017dynamic}
S.~Sabour, N.~Frosst, G.~E. Hinton, Dynamic routing between capsules, in:
  Advances in neural information processing systems, 2017, pp. 3856--3866.

\bibitem{lecun1998gradient}
Y.~LeCun, L.~Bottou, Y.~Bengio, P.~Haffner, Gradient-based learning applied to
  document recognition, Proceedings of the IEEE 86~(11) (1998) 2278--2324.

\bibitem{nair2010rectified}
V.~Nair, G.~E. Hinton, Rectified linear units improve restricted boltzmann
  machines, in: Proceedings of the 27th international conference on machine
  learning (ICML-10), 2010, pp. 807--814.

\bibitem{hochreiter1991untersuchungen}
S.~Hochreiter, Untersuchungen zu dynamischen neuronalen netzen, Diploma,
  Technische Universit{\"a}t M{\"u}nchen 91~(1).

\bibitem{nagi2011max}
J.~Nagi, F.~Ducatelle, G.~A. Di~Caro, D.~Cire{\c{s}}an, U.~Meier, A.~Giusti,
  F.~Nagi, J.~Schmidhuber, L.~M. Gambardella, Max-pooling convolutional neural
  networks for vision-based hand gesture recognition, in: 2011 IEEE
  International Conference on Signal and Image Processing Applications
  (ICSIPA), IEEE, 2011, pp. 342--347.

\bibitem{zou2019primer}
J.~Zou, M.~Huss, A.~Abid, P.~Mohammadi, A.~Torkamani, A.~Telenti, A primer on
  deep learning in genomics, Nature genetics 51~(1) (2019) 12--18.

\bibitem{lanchantin2017deep}
J.~Lanchantin, R.~Singh, B.~Wang, Y.~Qi, Deep motif dashboard: Visualizing and
  understanding genomic sequences using deep neural networks, in: Pacific
  Symposium on Biocomputing 2017, World Scientific, 2017, pp. 254--265.

\bibitem{shen2018recurrent}
Z.~Shen, W.~Bao, D.-S. Huang, Recurrent neural network for predicting
  transcription factor binding sites, Scientific reports 8~(1) (2018) 1--10.

\bibitem{lu2019predicting}
W.~Lu, Y.~Tang, H.~Wu, H.~Huang, Q.~Fu, J.~Qiu, H.~Li, Predicting rna secondary
  structure via adaptive deep recurrent neural networks with energy-based
  filter, BMC bioinformatics 20~(25) (2019) 1--10.

\bibitem{nguyen2019iprodna}
B.~P. Nguyen, Q.~H. Nguyen, G.-N. Doan-Ngoc, T.-H. Nguyen-Vo, S.~Rahardja,
  iprodna-capsnet: identifying protein-dna binding residues using capsule
  neural networks, BMC bioinformatics 20~(23) (2019) 1--12.

\bibitem{altschul1997gapped}
S.~F. Altschul, T.~L. Madden, A.~A. Sch{\"a}ffer, J.~Zhang, Z.~Zhang,
  W.~Miller, D.~J. Lipman, Gapped blast and psi-blast: a new generation of
  protein database search programs, Nucleic acids research 25~(17) (1997)
  3389--3402.

\bibitem{bairoch2000swiss}
A.~Bairoch, R.~Apweiler, The swiss-prot protein sequence database and its
  supplement trembl in 2000, Nucleic acids research 28~(1) (2000) 45--48.

\bibitem{du2020capsnet}
W.~Du, Y.~Sun, G.~Li, H.~Cao, R.~Pang, Y.~Li, Capsnet-ssp: multilane capsule
  network for predicting human saliva-secretory proteins, BMC Bioinformatics
  21~(1) (2020) 1--17.

\bibitem{song2020ac}
J.~Song, S.~Tian, L.~Yu, Y.~Xing, Q.~Yang, X.~Duan, Q.~Dai, Ac-caps: Attention
  based capsule network for predicting rbp binding sites of lncrna,
  Interdisciplinary Sciences: Computational Life Sciences (2020) 1--10.

\bibitem{umarov2019promoter}
R.~Umarov, H.~Kuwahara, Y.~Li, X.~Gao, V.~Solovyev, Promoter analysis and
  prediction in the human genome using sequence-based deep learning models,
  Bioinformatics 35~(16) (2019) 2730--2737.

\bibitem{schneider1990sequence}
T.~D. Schneider, R.~M. Stephens, Sequence logos: a new way to display consensus
  sequences, Nucleic acids research 18~(20) (1990) 6097--6100.

\bibitem{levy2020investigation}
J.~Levy, Y.~Chen, N.~Azizgolshani, C.~L. Petersen, A.~J. Titus, E.~L. Moen,
  L.~J. Vaickus, L.~A. Salas, B.~Christensen, Investigation of capsule-inspired
  neural network approaches for dna methylation, bioRxiv.

\end{thebibliography}

\end{document}